\documentclass[letterpaper]{article} 
\usepackage{aaai24}  
\usepackage{times}  
\usepackage{helvet}  
\usepackage{courier}  
\usepackage[hyphens]{url}  
\usepackage{graphicx} 
\usepackage{natbib}  
\usepackage{caption} 
\usepackage{algorithm}
\usepackage{algorithmic}

\usepackage{amsmath}
\usepackage{diagbox}
\usepackage{colortbl}
\usepackage{comment}
\usepackage{bm}
\usepackage{enumitem}

\usepackage{amssymb}
\usepackage{multirow}
\usepackage{booktabs}
\usepackage{caption,subcaption}
\usepackage{graphicx}
\usepackage{xcolor} 
\usepackage{makecell} 

\usepackage{newfloat}
\usepackage{listings}
\DeclareCaptionStyle{ruled}{labelfont=normalfont,labelsep=colon,strut=off} 
\floatstyle{ruled}
\newfloat{listing}{tb}{lst}{}
\floatname{listing}{Listing}
\title{Dirichlet-Based Prediction Calibration for Learning with Noisy Labels}
\author{
Chen-Chen Zong, Ye-Wen Wang, Ming-Kun Xie, Sheng-Jun Huang\thanks{Corresponding author.}
}
\affiliations{
College of Computer Science and Technology/Artificial Intelligence, Nanjing University of Aeronautics and Astronautics\\
MIIT Key Laboratory of Pattern Analysis and Machine Intelligence, Nanjing, China\\
\{chencz, linuswangg, mkxie, huangsj\}@nuaa.edu.cn

}

\usepackage{bibentry}

\begin{document}

\maketitle

\begin{abstract}
Learning with noisy labels can significantly hinder the generalization performance of deep neural networks (DNNs). Existing approaches address this issue through loss correction or example selection methods. However, these methods often rely on the model's predictions obtained from the softmax function, which can be over-confident and unreliable. In this study, we identify the translation invariance of the softmax function as the underlying cause of this problem and propose the \textit{Dirichlet-based Prediction Calibration} (DPC) method as a solution. Our method introduces a calibrated softmax function that breaks the translation invariance by incorporating a suitable constant in the exponent term, enabling more reliable model predictions. To ensure stable model training, we leverage a Dirichlet distribution to assign probabilities to predicted labels and introduce a novel evidence deep learning (EDL) loss. The proposed loss function encourages positive and sufficiently large logits for the given label, while penalizing negative and small logits for other labels, leading to more distinct logits and facilitating better example selection based on a large-margin criterion. Through extensive experiments on diverse benchmark datasets, we demonstrate that DPC achieves state-of-the-art performance. The code is available at https://github.com/chenchenzong/DPC.
\end{abstract}

\section{Introduction}

Large-scale datasets with high-quality annotations are crucial for achieving remarkable performance in deep neural networks (DNNs). However, collecting a large number of accurately annotated data is often costly and time-consuming. Recently, crowdsourcing labeling has become a mainstream solution to this problem due to its cost-effectiveness \citep{hossain2015crowdsourcing,chen2014gmission,huang2021asynchronous}. While the labeling cost is significantly reduced, it often introduces noisy labels unavoidably due to the various levels of expertise possessed by different labelers. Directly learning with such noisy labels can easily degrade the generalization performance of DNNs \citep{zhang2021understanding}. Therefore, training robust DNNs with noisy labels has become a challenge that attracted significant attention in recent years \citep{han2018co,huang2019o2u,li2020dividemix,liu2020early,zong2022noise,karim2022unicon}. 

Existing methods can be broadly classified into two categories: loss correction \citep{patrini2017making,ma2018dimensionality,zhang2018generalized,shu2019meta,ma2020normalized} and example selection \citep{garcia2016noise,malach2017decoupling,huang2019o2u,li2020dividemix,zhou2021robust,karim2022unicon}. The former aims to correct the loss by estimating the noise transition matrix, adjusting the example labels or weights. The latter attempts to separate clean examples from noisy ones based on the small-loss criterion \citep{li2020dividemix}, where the examples with low loss are assumed to have clean labels, and further consider recognized mislabeled examples as unlabeled ones to perform semi-supervised learning. Despite these two kinds of methods making great progress in dealing with noisy labels, most of them often suffer from the unreliability issue since standard DNNs can easily produce over-confident but incorrect predictions \citep{sensoy2018evidential,wang2021rethinking,xie2023dirichlet}. For example, to demonstrate the over-confidence issue of DNNs, \citet{sensoy2018evidential} experimented on handwritten digit recognition and observed that when the digit ``1" is rotated at an angle greater than 60 degrees, the model tends to output the digit ``2" with very high confidence.  

\begin{figure*}
	\centering
	\includegraphics[width=0.76\textwidth]{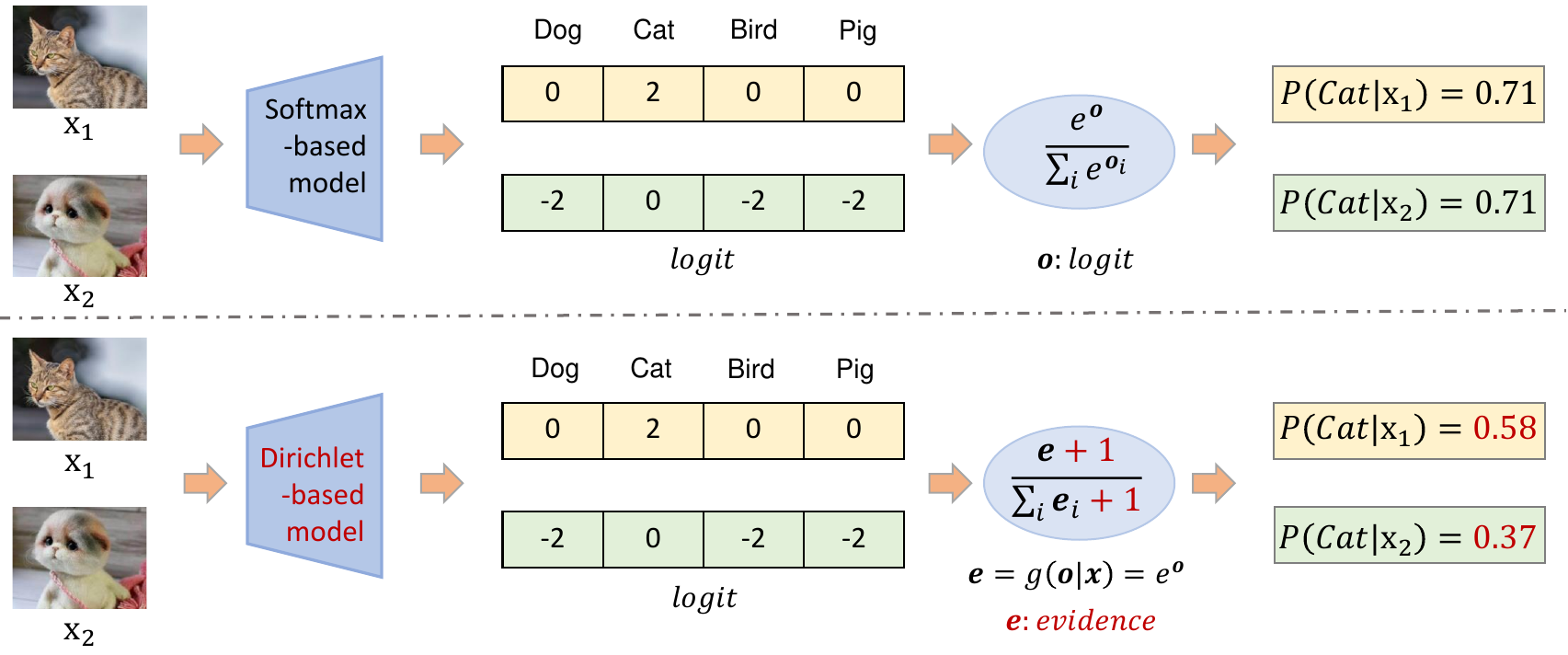}
 \caption{A specific case comparing the softmax-based model and our proposed calibrated Dirichlet-based model. The softmax function has translation invariance, \textit{i.e.}, can only reflect the relative relationship
between logits, and gives the same prediction for $\boldsymbol{\mathrm {x}}  _1 $ and $\boldsymbol{\mathrm {x}}  _2$, which contradicts our subjective intuition. We break the translation invariance by placing a suitable constant on the exponent term and proposing a corresponding Dirichlet-based training method.}
	\label{fig.1}
\end{figure*}

In this paper, we first disclose that the occurrence of the over-confidence phenomenon is caused by the translation invariance of the softmax function, \textit{i.e.}, adding or subtracting a constant from all logits does not alter the softmax values, and this actually elevates the probability of misclassifying mislabeled examples as clean ones in noisy label learning. Based on these findings, we develop a Dirichlet-based Prediction Calibration (DPC) method to calibrate the softmax probabilities and thus solve the over-confidence issue. Specifically, to obtain reliable confidence, a suitable constant is added to the exponent term of the softmax function to break its translation invariance. To solve the gradient shrinking issue caused by the calibration, a Dirichlet training scheme is developed to enforce the logits on the given label and other labels to be as separated as possible. With more distinguishable logits, we further design a large-margin criterion to achieve better performance on example selection.

The main contributions are summarized as follows:
\begin{itemize}
\item We disclose that softmax's translation invariance leads to over-confidence problems and may amplify the risk of misclassifying mislabeled examples as clean ones.

\item We propose a Dirichlet-based prediction calibration method to overcome over-confidence problems. The method calibrates the softmax function by breaking its translation invariance and improves the distinctiveness of predicted logits by designing an EDL loss. 

\item We drive a large-margin example selection criterion that is more suitable to the calibrated model. Compared to the small-loss criterion, our large-margin criterion is able to take full advantage of the distinguishable logits and thus achieve better example selection.

\item We conduct extensive experiments on benchmark and real-world datasets to demonstrate that DPC can achieve competitive performance compared with state-of-the-art methods.
\end{itemize}

\section{Related Work}




\textbf{Loss Correction.} This category of methods is implemented mainly in three ways: noise transition estimation, label correction, and example reweighting. For the first type, \citet{goldberger2017training} estimated the noise transition matrix by adding an additional linear layer on the top of the neural networks to correct the loss function. \citet{patrini2017making} proposed to exploit anchor points (data points that belong to a specific class almost surely) to obtain a pre-calculated Backward or Forward noise transition matrix. Label correction aims to correct the noisy labels. \citet{reed2014training} first proposed to update example labels by using their pseudo-labels in each training epoch. \citet{huang2020self} introduced the exponential moving average into the label refurbishment process to alleviate the instability issue caused by the instantaneous prediction. Example reweighting tries to eliminate the effects of noise labels by assigning small weights to the possibly mislabeled examples. \citet{chang2017active} used the example prediction variance as its weight to emphasize examples with inconsistent predictions. \citet{shu2019meta} introduced meta-learning to automatically learn an explicit loss-weight function based on an additional clean dataset.

\textbf{Example Selection.} This category of methods attempts to directly identify potentially noisy examples and then learn only based on clean examples or learning in a semi-supervised manner. \citet{huang2019o2u} proposed a straightforward noisy label detection approach named O2U-net, which requires adjusting the learning rate to keep the network transferring from overfitting to underfitting cyclically and then distinguishes examples by their accumulated loss values. \citet{jiang2018mentornet} firstly trains a teacher network and then uses it to select small loss examples as clean examples for guiding the training of the student network. Co-teaching \citep{han2018co} and Co-teaching+ \citep{yu2019does} maintain two networks simultaneously and let them select training examples for each other. DivideMix \citep{li2020dividemix} leverages the Gaussian mixture model to distinguish clean and mislabeled data and introduces a semi-supervised technique called MixMatch \citep{berthelot2019mixmatch} to leverage recognized mislabeled examples. \citet{liu2020early} and \citet{bai2021understanding} further improve this by proposing an early learning regularization term and a progressive early stopping technique to prevent the model from memorization noisy labels, respectively.

\section{Analysis of the Over-Confidence Phenomenon}

\begin{figure*}
	\centering
	\begin{subfigure}{0.24\linewidth}
		\includegraphics[width=1.\linewidth]{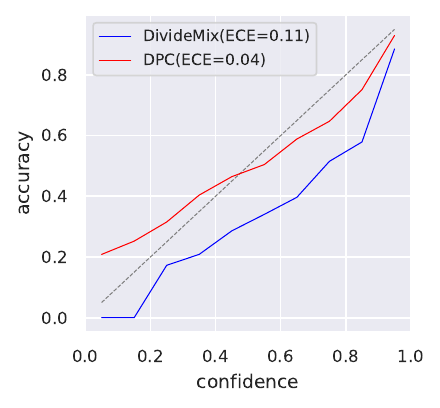}
		\caption{}
		\label{fig.2a}
	\end{subfigure}
        \hfill
        \begin{subfigure}{0.247\linewidth}
		\includegraphics[width=1.\linewidth]{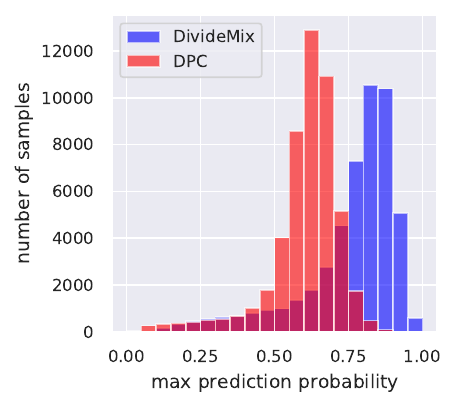}
		\caption{}
		\label{fig.2b}
	\end{subfigure}
	\hfill
	\begin{subfigure}{0.247\linewidth}
		\includegraphics[width=1.\linewidth]{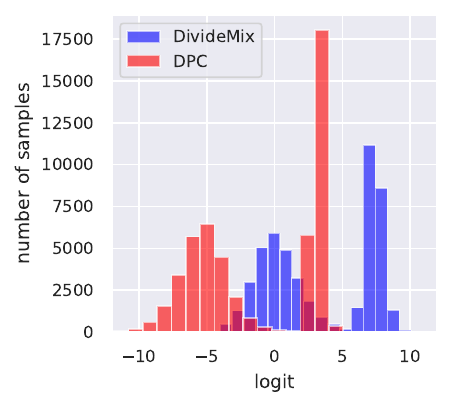}
		\caption{}
		\label{fig.2c}
	\end{subfigure}
        \hfill
	\begin{subfigure}{0.247\linewidth}
		\includegraphics[width=1.\linewidth]{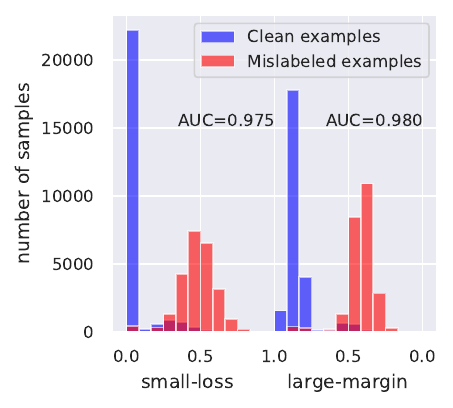}
		\caption{}
		\label{fig.2d}
	\end{subfigure}
	\caption{Training on CIFAR-10 with a 50\% symmetric noise rate. All the figures are plotted based on the results of the last epoch. (a) The Expected Calibration Error (ECE) results of the test data. A smaller ECE value is better, and correspondingly, a line closer to the dashed line is preferred. We can see that the softmax-based DivideMix tends to produce over-confident predictions. (b) The distribution of the maximum predicted probability for training examples. (c) The distribution of the given label logit for training examples. (d) The comparison of the example selection criterion. We can see that the proposed large-margin criterion can produce more discriminative results.}
	\label{fig.2}
\end{figure*}

Formally, let $\mathcal{D} =\left \{ \left ( \boldsymbol{x}_i,\boldsymbol{y}_i \right )  \right \} ^N_{i=1}$ denotes the corrupted training set with $N$ training examples, where each $\boldsymbol{x}_i$ represents an example and $\boldsymbol{y}_i\in \left \{  0,1\right \} ^C$ is the corrupted one-hot label over $C$ classes. Given a model $f(\cdot ;\boldsymbol{\theta })$ parameterized by $\boldsymbol{\theta }$, we fit the training data by minimizing the cross entropy loss:
\[\mathcal{L}_{ce}=\frac{-1}{N} \sum_{i=1}^{N} \boldsymbol{y}_i \log_{}{\boldsymbol{\rho}_i} =\frac{-1}{N} \sum_{i=1}^{N}\sum_{c=1}^{C}  y_{ic}\log_{}{\frac{e^{o_{ic}}}{ {\textstyle \sum_{j=1}^{C}e^{o_{ij}}  } } },\]
where $\boldsymbol{o}_i = f(\boldsymbol{x}_i;\boldsymbol{\theta }) =\left [ o_{i1},o_{i2},\dots,o_{iC} \right ] $ denotes the logit vector, $\boldsymbol{\rho}_i =\left [ \rho_{i1},\rho_{i2},\dots,\rho_{iC} \right ] $ is the softmax predicted probability vector by applying the softmax function to $\boldsymbol{o}_i$.

Our goal is to train a sufficiently good model based on the corrupted dataset $\mathcal{D}$. Towards this goal, many current methods \citep{li2020dividemix,bai2021understanding,karim2022unicon} first partitioned the whole dataset into clean and mislabeled subsets and then performing semi-supervised learning by treating the former as labeled, while the latter as unlabeled. For example, DivideMix \citep{li2020dividemix}, as a representative method, fits a two-component Gaussian Mixture Model (GMM) \citep{permuter2006study} to the example loss and uses the posterior probability of the component with a smaller mean to partition clean and mislabeled subsets. Then, a semi-supervised learning method called MixMatch \citep{berthelot2019mixmatch} is adopted.

It is noteworthy that both example selection and label correction highly depend on the quality of model predictions. If the predicted probabilities are closer to true ones, then the model would select clean examples more precisely. We perform an experiment to show that the model trained by DivideMix often suffers from the over-confidence issue based on Expected Calibration Error (ECE) \citep{guo2017calibration}, which is the most commonly used metric to quantify how well a deep learning model's predicted probabilities align with the actual probabilities of the events it predicts. Figure \ref{fig.2a} illustrates the ECE results of DivideMix on the test data of CIFAR-10 with 50\% symmetric noise. We can see that the confidence of the DivideMix model is much higher than the accuracy, \textit{i.e.}, the model becomes overconfident and produces unreliable predicted probability. 

To illustrate the intuition behind the phenomenon, we provide a specific case in Figure \ref{fig.1} where examples $\boldsymbol{\mathrm {x}} _1 $ with logits [0, 2, 0, 0] and $\boldsymbol{\mathrm {x}}  _2$ with logits [-2, 0, -2, -2] have equal probabilities of being predicted as ``cat". Since the softmax function can only reflect the relative relationship among logits of different classes, it would not change the softmax value by adding/subtracting a constant from all logits. This ``either-or'' prediction strategy does not exhibit errors on common closed-set datasets with clean labels. However, it becomes unreliable when confronted with abnormal examples and may yield erroneous results. For example, $\boldsymbol{\mathrm {x}} _2$ lacks typical ``cat" characteristics compared to $\boldsymbol{\mathrm {x}} _1$; still, it receives a high-confidence ``cat" label due to its dissimilarity to other categories. In essence, the model's confidence in $\boldsymbol{\mathrm {x}} _2$ isn't based on its clear ``cat"-like features. In noisy label learning, we tend to identify the most likely clean examples and leverage them to assist in exploiting all the available data. Obviously, $\boldsymbol{\mathrm {x}}  _1$, with unmistakable ``cat" features, is more likely to be a clean example and would be expected to have a larger probability than $\boldsymbol{\mathrm {x}}  _2 $. This is also supported by Dempster–Shafer Theory of Evidence (DST) \citep{dempster1968generalization} that a smaller/larger logit means that there is less/more evidence to support belonging to that class. 

Based on the preceding findings, we can conclude that the translation invariance of the softmax function would lead to the over-confidence phenomenon in noisy label learning, \textit{e.g.}, potential mislabeled examples with logits like $\boldsymbol{\mathrm {x}} _2$ may end up with probabilities equal to those of clean examples with logits like $\boldsymbol{\mathrm {x}} _1$, despite their logits being significantly smaller than $\boldsymbol{\mathrm {x}} _1$. This makes it challenging for the model to effectively perform label correction or example selection.

\section{The Proposed Method}

 To meet the challenge, we first propose a \textit{Dirichlet-based Prediction Calibration} (DPC) approach. As shown in Figure \ref{fig.1}, by placing a suitable constant on the exponent term of the softmax function, we can easily break softmax's translation invariance and calibrate the model prediction. Compared to DivideMix, DPC can significantly reduce ECE, \textit{i.e.} the red line is closer to the dashed line in Figure \ref{fig.2a}, and avoid producing over-confident model predictions (Figure \ref{fig.2b}). Meanwhile, we can see in Figure \ref{fig.2c} that DPC gives a more specific meaning to the logit, where a logit greater/less than 0 indicates that there is evidence/no evidence for the example belonging to that class. Therefore, we drive a large-margin criterion for example partitioning. In Figure \ref{fig.2d}, the AUC value of the large-margin criterion is larger than the small-loss criterion and achieves better distinguishable results.

\subsection{Dirichlet-Based Prediction Calibration}
As mentioned above, the softmax operator that converts the logit vector $\boldsymbol{o}$ to the probability vector $\boldsymbol{\rho}$ often leads the model predictions to be over-confident. To obtain more reliable output probabilities, we propose a calibrated softmax function and express the predicted probability for $\boldsymbol{x}_i$ as:
\begin{equation}\label{eq2}
\hat{\rho}_{ic}=\frac{e^{o_{ic}}+\gamma  }{ {\textstyle \sum_{j=1}^{C}\left ( e^{o_{ij}}+\gamma  \right ) } }, \quad c=1,2,\dots ,C,
\end{equation}
where $\gamma $ is a constant. This simple transformation can break the translation invariance of the softmax function and can significantly alleviate the over-confidence issue. However, since the gradient difference before and after calibration is constantly greater than 0 on the complementary labels (labels other than the given label), the calibration leads the commonly used cross-entropy loss to suffer from the gradient shrinking issue, which can be demonstrated as:
\begin{equation}\label{eq3}
\begin{aligned}
&\frac{\partial \mathcal{L}_{ce|\boldsymbol{\rho}_i} }{\partial o_{ic}} |_{\boldsymbol{x}_i}-\frac{\partial \mathcal{L}_{ce|\boldsymbol{\hat{\rho}}_i} }{\partial o_{ic}} |_{\boldsymbol{x}_i}\\&=\frac{\gamma  Ce^{o_{ic}}}{ {\textstyle \sum_{j=1}^{C}e^{o_{ij}}\sum_{j=1}^{C}\left ( e^{o_{ij}}+\gamma   \right )  } }> 0,\; \forall y_{ic}=0 .
\end{aligned}
\end{equation}
This motivates us to design a specific loss to compress the probabilities of complementary labels. However, simply adding a regularization term to the model output (\textit{e.g.}, KL divergence or L2 regularization) provides excessively strong constraints and can not achieve this while ensuring optimal model performance. Evidential deep learning (EDL) \citep{sensoy2018evidential,xie2023dirichlet} regards the probability as a random variable and indirectly optimizes it by optimizing the parameters of the distribution, which is a softer constraint and can help us achieve this goal. Based on this, we propose a novel training strategy to seamlessly incorporate EDL with our calibrated softmax function. Below, we provide a detailed description of the training strategy.

EDL terms evidence $\boldsymbol{e}_i=g(\boldsymbol{o}_i )$ as a measure of the amount of support collected from data in favor of an example to be classified into a certain class, where $g(\cdot )$ is a function (\textit{e.g.} exponential function) to ensure non-negative $\boldsymbol{e}_i$. Unlike traditional DNNs which give a point estimate of $\boldsymbol{\rho}$, EDL regards $\boldsymbol{\rho}$ as a random variable and places a Dirichlet distribution over $\boldsymbol{\rho}$ to represent the probability density of each possible $\boldsymbol{\rho}$. Specifically, for a given example $\boldsymbol{x}_i$, the probability density function of $\boldsymbol{\rho}_i$ is denoted as:
\[\begin{aligned}
&p\left ( \boldsymbol{\rho}_i|\boldsymbol{x}_i,\boldsymbol{\theta } \right ) = Dir\left ( \boldsymbol{\rho}_i|\boldsymbol{\alpha}_i \right )\\&=\begin{cases}
 \frac{\Gamma \left (  {\textstyle \sum_{j=1}^{C}\alpha _{ij}}  \right ) }{\Pi _{j=1}^C\Gamma \left ( \alpha _{ij} \right ) } \Pi _{j=1}^C\rho_{ij}^{\alpha _{ic}-1}, & \text{ if } \boldsymbol{\rho}_i\in \bigtriangleup ^C, \\
0,  & \text{ otherwise },
\end{cases}
\end{aligned}\]
where $\boldsymbol{\alpha }_i$ denotes the parameters of the Dirichlet distribution for $\boldsymbol{x}_i$, $\Gamma (\cdot )$ is the Gamma function and $\bigtriangleup ^C =\left \{ \boldsymbol{\rho}_i\mid  {\textstyle \sum_{j=1}^{C}\rho_{ij}=1 \; \text{and} \; \forall j \; 0\le \rho_{ij}\le 1}   \right \} $ is a C-dimensional unit simplex. Here, we define $\boldsymbol{e }_i=\gamma \left ( \boldsymbol{\alpha }_i-1 \right ) $. By marginalizing over $\boldsymbol{\rho}_i$, we can obtain the predicted probability for a given class $c$ as:
\begin{equation}\label{eq5}
\begin{aligned}
P(y=c|\boldsymbol{x}_i,\boldsymbol{\theta })&=\int p(y=c|\boldsymbol{\rho}_i)p(\boldsymbol{\rho}_i|\boldsymbol{x}_i,\boldsymbol{\theta })d\boldsymbol{\rho}_i\\&=\frac{\alpha _{ic}}{ {\textstyle \sum_{j=1}^{C}\alpha _{ij}} } =\frac{g(o_{ic})+\gamma  }{\sum_{j=1}^{C}\left ( g(o_{ij})+\gamma  \right ) }  .
\end{aligned}
\end{equation}
 Specifically, we can bridge the connection with Equation \ref{eq2} by defining $g(\cdot)$ as an exponential function. Following that, we train the EDL model by driving it to produce a sharp Dirichlet distribution situated at the corner of $\bigtriangleup ^C$ for all labeled data. To ensure this, on one hand, we minimize the negative logarithm of the marginal likelihood ($\mathcal{L} _{nll}$) to ensure the correctness of prediction:
\begin{equation}\label{eq6}
\begin{aligned}
\mathcal{L} _{nll}&=-\frac{1}{N} \sum_{i=1}^{N} \log_{}{ \left [ P(y=c|\boldsymbol{x}_i,\boldsymbol{\theta }) \right ] } \\&=\frac{1}{N} \sum_{i=1}^{N}\sum_{c=1}^{C}y_{ic}\left [ \log_{}{ {\textstyle  {\textstyle \left ( \sum_{j=1}^{C}\alpha _{ij} \right )-\log_{}{\alpha _{ic}}  } } } \right ].  
\end{aligned}\end{equation}
On the other hand, to cope with the problem posed by Equation \ref{eq3} and regularize the predictive distribution, a KL-divergence term $\mathcal{L} _{kl}$ is adopted by penalizing the evidence of the complementary labels to approach 0:
\begin{equation}\label{eq7}
\begin{aligned}
\mathcal{ L}_{kl} &=\frac{1}{NC} \sum_{i=1}^{N} D_{KL}\left(Dir\left ( \boldsymbol{\rho}_i|\tilde{\boldsymbol{\alpha}}_i  \right ) \parallel Dir\left ( \boldsymbol{\rho}_i|\boldsymbol{1} \right )\right) \\& =\frac{1}{NC} \sum_{i=1}^{N} \left [ \log_{}{\left [ \frac{\Gamma \left (  {\textstyle \sum_{j=1}^{C}\tilde{\alpha} _{ij}}  \right ) }{\Gamma\left ( C \right ) \Pi _{j=1}^C\Gamma \left ( \tilde{\alpha} _{ij} \right ) } \right ] } \right. \\& \left. +\sum_{c=1}^{C} \left (  \tilde{\alpha}_{ic}-1 \right ) \left [ \psi \left (  \tilde{\alpha}_{ic}\right )-  \psi\left ( \sum_{j=1}^{C} \tilde{\alpha}_{ij} \right )  \right ]   \right ] ,
\end{aligned}
\end{equation}
where $\tilde{\boldsymbol{\alpha}}_i=\boldsymbol{y}_i+(1-\boldsymbol{y}_i)\odot \boldsymbol{\alpha}_i$ can be seen as the Dirichlet parameters after removal of the given label evidence from predicted parameters $\boldsymbol{\alpha}$, $\boldsymbol{1}$ is a vector consisting of C ones and $\psi \left ( \cdot  \right ) $ represents the digamma function. Then, the overall training loss can be formulated as:
\[\mathcal{L}_{edl} = \mathcal{L}_{nll} + \beta \mathcal{L}_{kl},\]
where $\beta$ is used to balance the two terms.

\subsection{Large-Margin Example Selection Criterion}



\begin{table*}[!t]
 \centering
 \small
 \begin{tabular}{l|c|c|c|c|c|c||c|c|c|c|c|c} 
  \toprule   
  \toprule 
     \multirow{3}{*}{\diagbox{Method}{Dataset}} &\multicolumn{6}{c||}{CIFAR-10}&\multicolumn{6}{c}{CIFAR-100}\\
     \cmidrule{2-13}
     & \multicolumn{3}{c|}{Symmetric} & \multicolumn{3}{c||}{Asymmetric} & \multicolumn{3}{c|}{Symmetric} & \multicolumn{3}{c}{Asymmetric}\\
     & 20\% & 50\%& 80\% & 10\% & 30\%& 40\%  & 20\% & 50\%& 80\% &10\% & 30\%& 40\% \\
   \midrule
   Cross-Entropy &88.8& 81.7& 76.1 &88.8& 81.7& 76.1  
   &61.8& 37.3& 8.8    &68.1& 53.3& 44.5\\
   Mixup \citep{zhang2017mixup} &93.3& 83.3&77.7   &93.3& 83.3&77.7    & 72.4&57.6 &48.1   & 72.4&57.6 &48.1\\ 
   PENCIL \citep{yi2019probabilistic} &92.4 & 89.1 &77.5 &93.1 &92.9 &91.6& 69.4 &57.5 &31.1 &76.0 & 59.3 &48.3\\
   JPL \citep{kim2021joint}& 93.5 & 90.2&35.7  & 94.2 & 92.5&90.7   &  70.9& 67.7&17.8 &  72.0& 68.1&59.5\\  
   DivideMix \citep{li2020dividemix}  & 95.7 &94.4  & 92.9   & 93.8 &92.5  & 91.7    &  76.9 & 74.2& 59.6  &  71.6 & 69.5& 55.1\\ 
   PES \citep{bai2021understanding}& 95.9 &95.1  & 93.1    &  - & -& - & 77.4 &74.3  & 61.6    &  - & -& - \\
   ELR+ \citep{liu2020early}& 95.8 &94.8  & 93.3  & 95.4 &\textbf{94.7}  & 93.0   &  77.6 & 73.6& 60.8 &  77.3 & 74.6& 73.2 \\
   MOIT+ \citep{ortego2021multi}&94.1&91.1&75.8&94.2&94.1& 93.2&75.9&70.1&51.4&77.4&75.1&74.0\\
   \rowcolor{gray!25} DPC& \textbf{96.1} &\textbf{95.2}  & \textbf{93.5}  & \textbf{95.5} &94.5  & \textbf{93.6}    &  \textbf{79.4} & \textbf{76.5}& \textbf{63.0}    &  \textbf{79.0} & \textbf{77.6}& \textbf{74.1}\\  
   \midrule 
   UniCon \citep{karim2022unicon}& 96.0 &95.6  & 93.9  & 95.3 &94.6  & 94.1    &  78.9 & 77.6& 63.9     &  78.2 & 75.6& 74.8\\ 
   \rowcolor{gray!25} DPC* & \textbf{96.5} &\textbf{95.9}  & \textbf{94.8}   & \textbf{95.4} &\textbf{95.3}  & \textbf{94.9}    &  \textbf{81.0} & \textbf{78.5}& \textbf{66.4}
  &  \textbf{80.5} & \textbf{79.7}& \textbf{75.6}\\
   
  \bottomrule
  \bottomrule
\end{tabular}
\caption
  {
   Test accuracy (\%) comparison with state-of-the-art methods on CIFAR-10 and CIFAR-100 with synthetic noise. For previous techniques, results are copied from their respective papers. For our method, results are reported over 3 random runs.
  }
 \label{tab2}
\end{table*}

To achieve the same predicted probability on a single example, the calibrated softmax needs to provide a logit distribution with greater differentiation than the commonly used softmax function. This leads the proposed calibration method to produce output logits distribution more distinguishable. Thus, we propose a large-margin example selection criterion and define the margin for a given example $\boldsymbol{x}_i$ as the difference of predicted logits between the given class and the largest probable classes of complementary labels:
\[margin\left ( \boldsymbol{x}_i \right ) =o_{ic}-\max{}_{j\ne c} o_{ij},\;\text{where}\; c=arg\max \boldsymbol{y}_i .\]
A larger margin yields that the model is more confident that the example belongs to the class $c$. Unlike the small-loss criterion, which requires a suitable loss function for example selection, our large-margin criterion only relates to the model itself without introducing any external information.

After obtaining margin values for all training examples, we fit a two-component GMM to model per-example margin distribution and use its prediction on the Gaussian component with a larger mean (larger margin) to divide examples.

\subsection{Combining with Semi-Supervised Learning}

Following the partitioning of the training set $\mathcal{D}$ into the clean subset $\mathcal{X}$ and the mislabeled subset $\mathcal{U}$, we adopt MixMatch \citep{berthelot2019mixmatch} as the semi-supervised learning framework for the subsequent training similar to \citep{li2020dividemix,karim2022unicon}. 

Specifically, for a pair of examples $\left ( \boldsymbol{x}_1,\boldsymbol{x}_2 \right ) $ in $\mathcal{X} \cup \mathcal{U} $ with their given labels $\left ( \boldsymbol{y}_1,\boldsymbol{y}_2 \right ) $ and model predictions $\left ( \boldsymbol{\rho}_1,\boldsymbol{\rho}_2 \right )$, we can obtain the mixed $\left ( \boldsymbol{x}',\boldsymbol{y}',\boldsymbol{\rho}'\right ) $ by:
\[\lambda \sim Beta\left ( \alpha  \right ),
\lambda'=\max \left ( \lambda ,1-\lambda  \right ) ,\]
\[\boldsymbol{t}'=\lambda' \boldsymbol{t}_1+(1-\lambda' )\boldsymbol{t}_2, \;\text{where}\; \boldsymbol{t}\in \left \{  \boldsymbol{x}, \boldsymbol{y}, \boldsymbol{\rho}\right \}.\]
We can thus drive a mixed clean set $\mathcal{X} '$ and a mixed mislabeled set $\mathcal{U} '$. To deeply integrate our method with MixMatch, for a given example $\left ( \boldsymbol{x}',\boldsymbol{y}' \right ) $ in $\mathcal{X} '$, the supervised loss $\mathcal{L}_{sup}$ is defined as:
\[\mathcal{L}_{sup} \left ( \boldsymbol{x}',\boldsymbol{y}' \right ) =\lambda'\mathcal{L}_{edl} \left ( \boldsymbol{x}_1,\boldsymbol{{y}}_1 \right )+(1-\lambda' )\mathcal{L}_{edl} \left ( \boldsymbol{x}_2,\boldsymbol{{y}}_2 \right ).\]
The unsupervised loss $\mathcal{L}_{uns}$ for another given example $\left ( \boldsymbol{x}',\boldsymbol{\rho}' \right ) $ in $\mathcal{U} '$ is defined as:
\[\mathcal{L}_{uns}\left ( \boldsymbol{x}' ,\boldsymbol{\rho}'\right ) =\left \| \boldsymbol{\rho}' - \boldsymbol{{\rho}}(f(\boldsymbol{x}';\boldsymbol{\theta})) \right \|_2^2 .\]
Eventually, by using the balancing factor $\lambda_{uns}$ to control the strength of $\mathcal{L}_{uns}$, the overall loss can be expressed as:
\[\mathcal{L}_{total} = \textstyle \sum_{ \mathcal{X} '}^{} \mathcal{L}_{sup}(\boldsymbol{x}',\boldsymbol{y}') + \lambda_{uns} \sum_{ \mathcal{U} '}^{}\mathcal{L}_{uns}(\boldsymbol{x}',\boldsymbol{\rho}').\]

Note that previous works \citep{li2020dividemix,karim2022unicon} calculate the supervised loss directly based on $\boldsymbol{x}'$, while our supervised loss is indirectly obtained by combining $\boldsymbol{x}_1$ and $\boldsymbol{x}_2$. This can ensure that models give a positive and sufficiently large logit on the given label and penalize other labels with a negative and as small as possible logit. Considering that the mixed examples with unsupervised loss may break this rule and drive the model to provide smooth outputs, we adopt a two-head network architecture, where one head is allocated for supervised loss and the other for unsupervised loss. Detailed derivation of Equation \ref{eq3},\ref{eq5},\ref{eq6} and \ref{eq7} and more detailed implementations are available in the supplementary file.


\begin{table*}[!t]
 \centering
 \small
 \begin{tabular}{l|c|c|c|c|c||c} 
  \toprule  
  \toprule
     \multirow{2}{*}{\diagbox{Method}{Dataset}} &\multicolumn{5}{c||}{CIFAR-10}&\multicolumn{1}{c}{CIFAR-100}\\
     \cmidrule{2-7}
     & aggre & rand1 & rand2  & rand3 & worst & noisy100\\
   \midrule
   Cross-Entropy &87.77$\pm$0.38& 85.02$\pm$0.65& 86.46$\pm$1.79    &85.16$\pm$0.61& 77.69$\pm$1.55& 55.50$\pm$0.66\\
   GCE~\citep{zhang2018generalized} &87.85$\pm$0.70& 87.61$\pm$0.28& 87.70$\pm$0.56 &87.58$\pm$0.29 & 80.66$\pm$0.35& 56.73$\pm$0.30\\
   Co-teaching~\citep{han2018co} &91.20$\pm$0.13&90.33$\pm$0.13& 90.30$\pm$0.17 &90.15$\pm$0.18& 83.83$\pm$0.13& 60.37$\pm$0.27\\
   PES~\citep{bai2021understanding} &94.66$\pm$0.18& 95.06$\pm$0.15& 95.19$\pm$0.23    &95.22$\pm$0.13& 92.68$\pm$0.22& 70.36$\pm$0.33\\
   ELR+~\citep{liu2020early} &94.83$\pm$0.10& 94.43$\pm$0.41&94.20$\pm$0.24   & 94.34$\pm$0.22&91.09$\pm$1.60 &66.72$\pm$0.07\\
   CORES~\citep{cheng2020learning} &95.25$\pm$0.09& 94.45$\pm$0.14&94.88$\pm$0.31    & 94.74$\pm$0.03&91.66$\pm$0.09 &61.15$\pm$0.73\\
   DivideMix~\citep{li2020dividemix}  & 95.01$\pm$0.71 &95.16$\pm$0.19  & 95.23$\pm$0.07    &  95.21$\pm$0.14 & 92.56$\pm$0.42& 71.13$\pm$0.48 \\
   SOP~\citep{liu2022robust}& 95.61$\pm$0.13 & 95.28$\pm$0.13&95.31$\pm$0.10   &  95.39$\pm$0.11& 93.24$\pm$0.21&67.81$\pm$0.23\\       
   \midrule  
   \rowcolor{gray!25} DPC& \textbf{95.77$\pm$0.23} &\textbf{95.97$\pm$0.07}  & \textbf{95.92$\pm$0.14}    &  \textbf{95.90$\pm$0.09} & \textbf{93.82$\pm$0.31}& \textbf{71.42$\pm$0.23} \\  
   
  \bottomrule
  \bottomrule
\end{tabular}
\caption
  {
   Comparison with state-of-the-art methods in test accuracy (\%) on CIFAR-N. The corresponding noise rate is ``aggre" (9.03\%), ``rand1" (17.23\%), ``rand2" (18.12\%), ``rand3" (17.64\%), ``worst" (40.21\%) and ``noisy100" (40.20\%). The results of comparing methods are copied from \citep{wei2021learning}. The results (mean$\pm$std) of our method are reported over 5 random runs.
  }
 \label{tab3}
\end{table*}

\section{Experiments}


\textbf{Datasets.} We experimentally demonstrate the effectiveness of DPC on both synthetic noise datasets (CIFAR-10 and CIFAR-100 \citep{krizhevsky2009learning}) and real-world noise datasets (CIFAR-10N, CIFAR-100N \citep{wei2021learning} and WebVision \citep{li2017webvision}). All the CIFAR datasets contain 50K training images and 10K test images of size $32\times 32$. For CIFAR-10 and CIFAR-100, we manually inject two types of label noise: symmetric and asymmetric noise,  where the noise rate is set to 20\%, 50\%, and 80\% for symmetric noise and 10\%, 30\%, and 40\% for asymmetric noise. CIFAR-10/100N is a re-annotation version of CIFAR-10/100 by human workers. Specifically, each image in CIFAR-10N owns three submitted labels denoted as ``rand1", ``rand2", and ``rand3", and two ensembled labels denoted as ``aggre", and ``worst". While in CIFAR-100N, each image only contains a single submitted label represented as ``noisy100". WebVision comprises 2.4 million images obtained from web crawling using 1K concepts included in ImageNet ILSVRC12. Here, following the previous studies \citep{li2020dividemix,karim2022unicon}, we only use the first 50 classes of the Google image subset to construct the training set.


\textbf{Training Details.} For CIFAR-10 and CIFAR-100, we use PreAct ResNet18 \citep{he2016identity} as the base model and train it by stochastic gradient descent (SGD) optimizer with momentum 0.9, weight decay 0.0005, and batch size 128 for 300 epochs. The initial learning rate is set to 0.02 and reduced by a factor of 10 after 150 epochs. The warm-up epoch is 10 for CIFAR-10 and 30 for CIFAR-100. For CIFAR-10N and CIFAR-100N, ResNet34 \citep{he2016deep} is adopted and the warm-up epoch is changed to 30 for CIFAR-10 and 40 for CIFAR-100, respectively. For WebVision, we train InceptionResNetV2 \citep{szegedy2017inception} from scratch with changed parameters of batch size 32, warm-up epoch 1, and training epoch 100. The initial learning rate is changed to 0.01 and reduced by a factor of 10 after 50 epochs. For all experiments, we set $\beta$ as $0.5$ and have $\gamma =\frac{10}{C} $. 

Note that methods employing stronger data augmentation techniques or integrating advanced technologies like self-supervision consistently push the limits of state-of-the-art performance. Our approach, however, operates orthogonally with these advanced techniques and is expected to yield improved performance in combined with them. In this paper, we mainly compare with state-of-the-art methods not using these techniques. Additionally, we compare with UniCon \citep{karim2022unicon} to verify the effectiveness of DPC when adopting strong data augmentation for training. For a fair comparison, we apply a consistency regularization term to implement the same strong data augmentation for our method (denoted as DPC*). More detailed implementations can be found in the supplementary file.

\subsection{Experimental Results}

Table \ref{tab2} shows the averaged test accuracy over the last 10 epochs on CIFAR-10 and CIFAR-100 with different levels of symmetric and asymmetric label noise. For methods without strong augmentation, DPC achieves the best performance in most cases among all the compared methods. Especially for CIFAR-100 with symmetric noise 20\% and 50\%, DPC gains at least 1.8\% performance improvement. For methods with strong augmentation, we can see that DPC* still outperforms UniCon by a large margin across all noise ratios. These experiments not only confirm that the over-confidence phenomenon of the model is detrimental to the task of learning with noisy labels, but also demonstrate the effectiveness of our proposed method.

Table \ref{tab3} reports the final round accuracy on CIFAR-10N and CIFAR-100N with realistic label noise. Although the model architecture is changed in this setting, our method still maintains a performance gain over all comparing methods.

\begin{table}[ht]
 \centering
 \small
 \begin{tabular} {l |c|c|c|c }
  \toprule  
  \toprule
   \multirow{2}{*}{Method}  & \multicolumn{2}{c|}{WebVision} & \multicolumn{2}{c}{ILSVRC12}\\
   \cmidrule{2-5}
   & top1 & top5& top1 & top5\\
   \midrule   
      
    D2L~\citep{ma2018dimensionality} &62.7 &84.0&  57.8 &81.4\\
   MentorNet~\citep{jiang2018mentornet}  &63.0 &81.4&  57.8 &79.9\\ 
   
   Co-teaching~\citep{han2018co}&63.6 &85.2&   61.5 &84.7\\   
   ICV~\citep{chen2019understanding}  &  65.2 &85.3&  61.6 &85.0\\ 
   ELR+ &77.8 &91.7& 70.3 &89.8\\
   DivideMix &77.3 &91.6& 75.2 &90.8\\
   \rowcolor{gray!25}DPC &\textbf{79.2} &\textbf{93.0}& \textbf{75.8} &\textbf{92.5}\\
   \midrule
   UniCon &77.6 &93.4& 75.3 &93.7\\
   \rowcolor{gray!25}DPC* &\textbf{81.1} &\textbf{93.5}& \textbf{78.0} &\textbf{93.8}\\
  \bottomrule
  \bottomrule
 \end{tabular}
 \caption
  {
  Accuracy comparison on the WebVision validation set and the ImageNet ILSVRC12 validation set.
  }
 \label{tab4}
\end{table}

\begin{table*}[!t]
 \centering
 \small
 \begin{tabular}{l|cc|cc|cc||cc|cc|cc} 
  \toprule  
  \toprule
     Dataset &\multicolumn{6}{c||}{CIFAR-10}&\multicolumn{6}{c}{CIFAR-100}\\
     \midrule
    Noise Rate & \multicolumn{2}{c|}{20\%} & \multicolumn{2}{c|}{50\%}& \multicolumn{2}{c||}{80\%}  &\multicolumn{2}{c|}{20\%} & \multicolumn{2}{c|}{50\%}& \multicolumn{2}{c}{80\%} \\
   \midrule
   Method &Best& Last& Best    &Last& Best& Last&Best& Last& Best    &Last& Best& Last\\
   \midrule
   DPC w/o $\mathcal{L}_{edl}$ &96.1& 96.0&94.9    & 94.7&93.3&93.0&79.4& 78.6& 75.6    &75.1& 60.3& 60.0\\    
   DPC w/o LM & 96.3 & 96.1&95.1   &  94.8& 93.3&93.1&77.8& 77.6& 76.5    &76.0& 57.4& 57.1\\  
   DPC w/o TH &96.3& 96.1&95.2    & 94.9&92.8&92.5&79.8& 79.6& 76.5    &76.1& 62.8& 62.4\\
   \midrule
   \rowcolor{gray!25}DPC & 96.3 &96.1 & 95.3   & 95.2 & 93.6& 93.5 &79.9& 79.4& 76.9   &76.5& 63.3& 63.0\\ 
   \midrule
   \midrule
   UniCon & 94.2 &93.4 & 95.6   & 94.9 & 94.2& 93.7 &79.0& 77.0& 77.1   &75.9& 65.0& 64.1\\ 
   UniCon w $\mathcal{L}_{edl}$ & 94.6 &93.8 & 95.5   & 94.8 & 94.2& 93.7 &79.5& 77.8& 77.5   &76.0& 65.4& 64.4\\ 
   \midrule
   \rowcolor{gray!25}Improve &$\uparrow$ 0.4 &$\uparrow$ 0.4 &$\downarrow$ 0.1&$\downarrow$ 0.1&$\uparrow$ 0.0 &$\uparrow$ 0.0 &$\uparrow$ 0.5 &$\uparrow$ 0.8& $\uparrow$ 0.4&$\uparrow$ 0.1&$\uparrow$ 0.4&$\uparrow$ 0.3\\
   \bottomrule
  \bottomrule
\end{tabular}
\caption
  {
   Ablation studies on CIFAR-10 and CIFAR-100 with symmetric noise. LM indices the large-margin example selection criterion. TH means a model with two classification heads. Here, we show the rerun results for UniCon.
  }
 \label{tab5}
\end{table*}

\begin{figure*}
	\centering
	\begin{subfigure}{0.245\linewidth}
		\includegraphics[width=1.\linewidth]{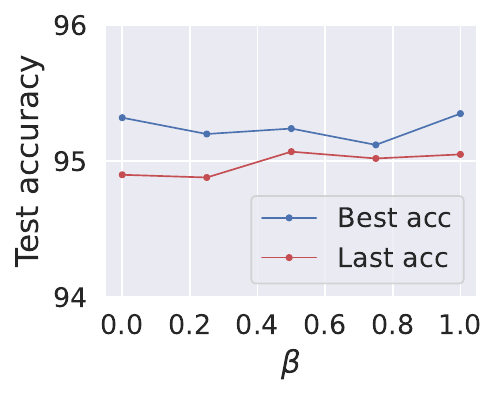}
		\caption{}
		\label{fig.3a}
	\end{subfigure}
        \hfill
        \begin{subfigure}{0.245\linewidth}
		\includegraphics[width=1.\linewidth]{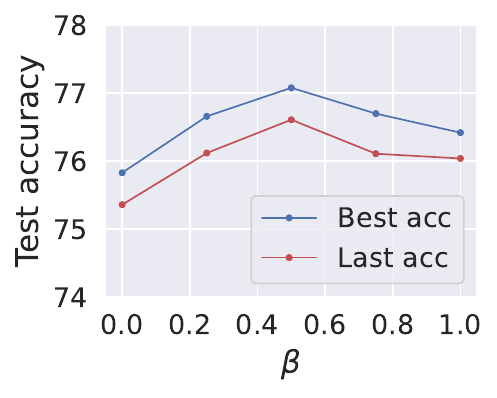}
		\caption{}
		\label{fig.3b}
	\end{subfigure}
	\hfill
	\begin{subfigure}{0.245\linewidth}
		\includegraphics[width=1.\linewidth]{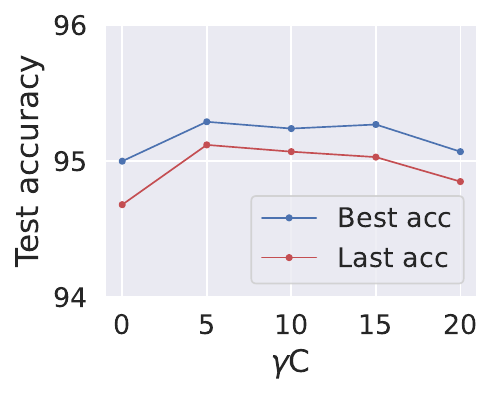}
		\caption{}
		\label{fig.3c}
	\end{subfigure}
        \hfill
	\begin{subfigure}{0.245\linewidth}
		\includegraphics[width=1.\linewidth]{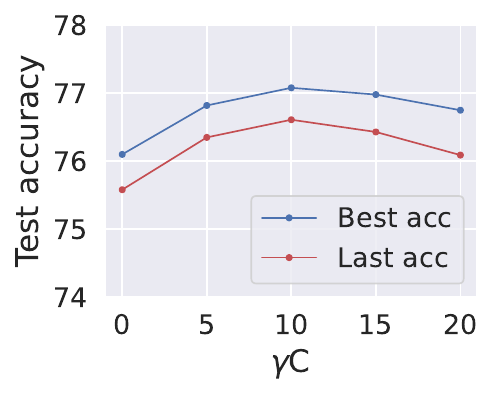}
		\caption{}
		\label{fig.3d}
	\end{subfigure}
	\caption{Ablation studies on CIFAR-10 and CIFAR-100 with a 50\% symmetric noise rate, respectively. (a) and (b) are the ablation studies of $\beta$ on CIFAR-10 and CIFAR-100. (c) and (d) are the ablation studies of $\gamma$ on CIFAR-10 and CIFAR-100.}
	\label{fig.3}
\end{figure*}

We further report the performance of our method on another real-world noise dataset WebVision. As shown in Table \ref{tab4}, DPC and DPC* both achieve the best results on the WebVision validation set and the ImageNet ILSVRC12 validation set. 

These experiments demonstrate that our method is equally effective for both synthetic and real-world label noise, and consistently performs well across different models.

\subsection{Ablation Studies}

\textbf{Effect of $\mathcal{L}_{edl}$.} Table \ref{tab5} indicates the impact of the proposed Dirichlet training scheme (denoted as $\mathcal{L}_{edl}$) on the overall performance of DPC. We can see that the performance of DPC without $\mathcal{L}_{edl}$ has a significant drop at various noise rates. This suggests that the softmax-based predicted probabilities do have bias, and it is necessary to calibrate existing noisy label learning methods. Furthermore, to verify the generalizability of the Dirichlet-based prediction calibration method, we also provide the results of UniCon integrated with $\mathcal{L}_{edl}$ in Table \ref{tab5}. As we can see, the proposed method still works effectively and has a certain degree of generality. Note that we have also provided integration results of two other methods with $\mathcal{L}_{edl}$ in the supplementary materials, and the results are consistent.

\textbf{Effect of Large-Margin Criterion.} DPC without large-margin criterion (denoted as LM), \textit{i.e.} with small-loss criterion, also shows unsatisfactory performance (see Table \ref{tab5}). This is because the two items in $\mathcal{L}_{edl}$ have different scales. For CIFAR-100 with noise ratio 80\%, $\mathcal{L}_{kl}$ is oscillating and provides unstable loss estimations. The proposed criterion can counter this since it is independent of a specific loss.

\textbf{Effect of Two Classification Heads.} The performance gap between one and two classification heads (denoted as TH) becomes large with the increase of noise rate as shown in Table \ref{tab5}. $\mathcal{L}_{uns}$ tends to make the model produce smooth logit vectors, while $\mathcal{L}_{sup}$ encourages the model to provide distinguishable logit vectors. From an example selection perspective, $\mathcal{L}_{uns}$ and $\mathcal{L}_{sup}$ are in conflict especially under high noise settings since a large noise rate commonly corresponds to a large $\lambda_{uns}$.

\textbf{Hyper-Parameter Sensitivity of $\beta$.} This parameter controls the tradeoff between $\mathcal{L}_{nll}$ and $\mathcal{L}_{kl}$. We test the sensitivity of $\beta$ on CIFAR-10 and CIFAR-100 with a 50\% symmetric noise rate, respectively. The results are shown in Figure \ref{fig.3a} and \ref{fig.3b}, where $\beta \in \left \{ 0,0.25,0.5,0.75,1 \right \} $. Since the CIFAR-10 task is relatively simple, the model is less sensitive to the value of $\beta$. However, for CIFAR-100, different values of $\beta$ have a significant impact on the accuracy. Finally, we determined the value of $\beta$ to be 0.5 and used this setting in all experiments.

\textbf{Hyper-Parameter Sensitivity of $\gamma$.} This parameter is the placed constant in the calibrated softmax function. Figure \ref{fig.3c} and \ref{fig.3d} present the results on CIFAR-10 and CIFAR-100 under 50\% symmetric noise rate, where $\gamma C \in \left \{ 0,5,10,15,20 \right \} $. According to the results, we recommend setting $\gamma C$ as $10$ for all experiments.

\section{Conclusion}

In this paper, we propose DPC to combat label noise by calibrating the predicted probability involved in the example selection and label correction procedure. DPC consists of two important components: Dirichlet-based prediction calibration and large-margin example selection criterion. The former includes a calibrated softmax function that can convert more accurate predicted probability from logit and a corresponding Dirichlet-based model training loss to ensure sufficient training. Through this, the model can avoid outputting over-confident predictions, and produce more distinguishable logit outputs which drive us to propose the latter, \textit{i.e.} the large-margin example selection criterion. Extensive experiments across multiple datasets demonstrate that our proposed method consistently exhibits substantial performance gains compared to the state-of-the-art methods.

\section{Acknowledgments}

This work was supported by the National Key R\&D Program of China (2020AAA0107000), the Natural Science Foundation of Jiangsu Province of China (BK20222012, BK20211517), and NSFC (62222605).

\bibliography{aaai24}

\end{document}